\documentclass{article} 
\usepackage{nips14submit_e,times}
\usepackage{hyperref}
\usepackage{url}

\usepackage{graphicx}
\usepackage{amsmath}
\usepackage{amssymb}
\usepackage{algorithmic}
\usepackage{algorithm}
\usepackage{subfigure}
\usepackage{array}
\usepackage{tabularx}
\usepackage{multirow}
\usepackage{pdfsync}
\usepackage[pagewise]{lineno}

\newcommand{\bfx}{{\textbf{x}}}

\newcommand{\bfw}{{\textbf{w}}}
\newcommand{\bfy}{{\textbf{y}}}

\title{Domain Transfer Structured Output Learning}

\author{
Jim Jing-Yan Wang\\
Computer, Electrical and Mathematical Sciences and Engineering Division\\
King Abdullah University of Science and Technology (KAUST)\\
Thuwal 23955-6900, Saudi Arabia\\
\texttt{jimjywang@gmail.com}
}

\nipsfinalcopy 

\begin{document}

\maketitle

\begin{abstract}
In this paper, we propose the problem of domain transfer structured output learning and the first solution to solve it. The problem is defined on two different data domains sharing the same input and output spaces, named as source domain and target domain. The outputs are structured, and for the data samples of the source domain, the corresponding outputs are available, while for most data samples of the target domain, the corresponding outputs are missing. The input distributions of the two domains are significantly different. The problem is to learn a predictor for the target domain to predict the structured outputs from the input. Due to the limited number of  outputs available for the samples form the target domain, it is difficult to directly learn the predictor from the target domain, thus it is necessary to use the output information available in source domain. We propose to learn the target domain predictor by adapting a auxiliary predictor trained by using source domain data to the target domain. The adaptation is implemented by adding a delta function on the basis of the auxiliary predictor. An algorithm is developed to learn the parameter of the delta function to minimize loss functions associated with the predicted outputs against the true outputs of the data samples with available outputs of the target domain.
\end{abstract}

\section{Introduction}

Recently, strutted learning has attracted much attention from both machine learning and computer vision communities \cite{tsochantaridis2004support}.
Traditional machine learning problems, including pattern classification and regression problems, try to learn a predictor to predict a discrete class label or a continuous response value from a input feature vector of a data sample. Different from the traditional machine learning, structured learning aims to learn predictors to predict structured outputs, such as vectors, sequences, trees, and graphs. Some structured learning methods have been developed for this purpose. For example, Tsochantaridis et al. \cite{tsochantaridis2004support} proposed the structured support vector machines (SSVM) proposed to generalize multiclass SVM learning by  involving features extracted jointly from inputs and structured outputs, and solving the resulting optimization problem with a cutting plane algorithm. Shen et al. \cite{shen2013structboost} standardized boosting approaches to structured learning, and proposed the a new boosting algorithm for structured output prediction, by formulating an equivalent 1-slack formulation and solving it using a combination of cutting planes and column generation.

A basic assumption of these methods is that for all the training samples, both the input feature vectors and the structured outputs are available for the learning problem. However, in the real-world applications, it may be expensive or difficult to obtain the structured output for a given data sample, making that the outputs of most training samples are unavailable. Learning from a limited number of training samples with corresponding structured outputs is usually not robust for a structured learning task. Recently, domain transfer learning \cite{duan2009domain} has been proposed to solve this problem with help of a data domain with adequate labeled data samples to learn a predictor for another domain with limited labels. The domain with adequate labeled data samples are named as source domain, while the other one is named as target domain. They share the same input and output space, but the input distributions are significantly different. Although lots of domain transfer learning algorithms have been developed, surprisingly, they are all limited to predict simple class label, and ignore the problem of predicting structured complex output. In this paper, we extend the domain transfer learning to predict structured outputs by defining a novel learning problem -- domain transfer structured output learning. This problem has the following features:

\begin{enumerate}
\item A source domain are provided to help learning a predictor for a target domain, and both the domains share the same input and output space.
\item The input distributions of source and target domains are significantly different, making it impossible to combine the two domains directly for the learning.
\item The outputs are structured and complex. The outputs of the data samples in the source domain are all available, while only a limited number of the target domain samples have corresponding outputs.
\end{enumerate}

To solve this problem, we first learn a structured output predictor using only the source domain, and then transfer it to the target domain by add a additional delta function. To learn the delta function parameter, we consider the data samples with outputs in the target domain with structured outputs. A objective function is build by comparing the predicted outputs against the true outputs of these data samples in the target domain, and is optimized using a cutting plane algorithm. The rest parts of this paper is organized as follows: In section \ref{sec:method}, we introduce the proposed method.
In section \ref{sec:conclusion}, the paper is concluded.

\section{Proposed Method}
\label{sec:method}

Given a source domain $\mathcal{S}$ with many input/output pairs, we can learn a scoring function $f^S(\bfx,y)$ to math a given input vector $\bfx\in \mathbb{R}^d$ and a structure output $y \mathcal{Y}$, and use it to construct a predictor to predict the output $y^*$ of a given input feature $\bfx$, $y_* = \max_{y\in \mathcal{Y}} f^S(\bfx,y)$. Given the target domain with $n$ data samples, denoted as $\mathcal{T}=\{(\bfx_1,y_1), \cdots, (\bfx_l,y_l), \bfx_{l+1}, \cdots, \bfx_n\}$, where only the first $l$ samples have outputs $\{y_i\}_{i=1}^l$, we try to learn a score function $f^T(\bfx,y)$ for it. It is defined as the combination of $f^S(\bfx,y)$ and a delta function $\Delta f(\bfx,y)$,
\begin{equation}
\begin{aligned}
f^T(\bfx,y) = f^S(\bfx,y) +\Delta f(\bfx,y)  = f^S(\bfx,y) + \bfw^\top \Psi(\bfx, y),
\end{aligned}
\end{equation}
where $\Psi(\bfx, y) = [\psi_1(\bfx,y),\cdots, \psi_m(\bfx,y)]^\top$ is a column vector of $m$ basis scoring functions, $\Delta f(\bfx,y) = \bfw^\top \Psi(\bfx, y)$ is defined as a linear combination of these basis functions, and $\bfw \in \mathbb{R}^m$ is the combination weight vector. To learn $\bfw$, we argue the following constrained minimization problem,

\begin{equation}
\begin{aligned}
\underset{\bfw,\xi}{\min}
~&
\left \{
\frac{1}{2} \| \bfw \|_2^2 + C \xi \right \}\\
s.t.~&
\forall \overline{\bfy}^k \in \mathcal{Y}^l: \\
&
\frac{1}{l} \sum_{i=1}^l \left [
\left(f^S(\bfx_i,y_i) + \bfw^\top \Psi(\bfx_i, y_i)\right) -
\left(f^S(\bfx_i,\overline{y}^k_i) + \bfw^\top \Psi(\bfx_i, \overline{y}^k_i)
\right)
\right ]
\geq
\frac{1}{l} \sum_{i=1}^l L (y_i,\overline{y}^k_i)
-\xi,
\xi \geq 0.
\end{aligned}
\end{equation}
where $\bfy_l = [y_1,\cdots,y_l]^\top$, $\overline{\bfy}^k = [\overline{y}^k_1,\cdots,\overline{y}^k_l]^\top$, and $L (y_i,\overline{y}_i)$ is a loss associated with a prediction $\overline{y}_i$ against the true structured output $y_i$. Please note that instead of minimizing $L (y_i,y^*_i)$ with $y^*_i = \max_{y\in \mathcal{Y}} f^T(\bfx,y)$, we minimize its upper boundary as in the constrain. The dual form of this problem is
\begin{equation}
\begin{aligned}
\underset{{\alpha_k}|_{k: \forall \overline{\bfy}^k \in \mathcal{Y}^l}}{\max}
~&
\left \{ -
\frac{1}{2}\sum_{k,k':\overline{\bfy}^k, \overline{\bfy}^{k'} \in \mathcal{Y}^l}
\alpha_k \left [ \frac{1}{l} \sum_{i=1}^l \left (\Psi(\bfx_i, y_i ) - \Psi(\bfx_i, \overline{y}^k_i) \right ) \right ]^\top
\left [ \frac{1}{l} \sum_{i=1}^l \left (\Psi(\bfx_i, y_i ) - \Psi(\bfx_i, \overline{y}^{k'}_i) \right ) \right ]
\alpha_{k'} \right .\\
& \left . + \sum_{k: \overline{\bfy}^k \in \mathcal{Y}^l} \alpha_k
\left [ L (y_i,\overline{y}^k_i) - \frac{1}{l} \sum_{i=1}^l \left (
f^S(\bfx_i,y_i)  + f^S(\bfx_i,\overline{y}^k_i) \right )
\right ]
\right \}\\
s.t.~&
\sum_{k: \overline{\bfy}^k \in \mathcal{Y}^l} \alpha_k \leq C,  \alpha_k \geq 0, k: \forall \overline{\bfy}^k \in \mathcal{Y}^l,
\end{aligned}
\end{equation}
where $\alpha_k$ is the Lagrange multiplier of the $k$-th constrain. This problem can be solved as a quadratic programming (QP) problem. After ${\alpha_k}_{k: \overline{\bfy}^k \in \mathcal{Y}^l} $ is solved, we can recover $\bfw$ as follows,
\begin{equation}
\begin{aligned}
\bfw=
\sum_{k: \overline{\bfy}^k \in \mathcal{Y}^l}
\alpha_k \left [ \frac{1}{l} \sum_{i=1}^l \left (\Psi(\bfx_i, y_i ) - \Psi(\bfx_i, \overline{y}^k_i) \right ) \right ]
\end{aligned}
\end{equation}
Based on this optimization result, we develop a cutting-plane algorithm to learn $\bfw$ as in Algorithm \ref{alg:iter}.

\begin{algorithm}[htb!]
\caption{Cutting-plane algorithm for solving delta function parameter $\bfw$.}
\label{alg:iter}
\begin{algorithmic}
\STATE \textbf{Input}: Source domain scoring function $f^S(\bfx,y)$;

\STATE \textbf{Input}: Target domain training set $\mathcal{T}=\{(\bfx_1,y_1), \cdots, (\bfx_l,y_l), \bfx_{l+1}, \cdots, \bfx_n\}$;

\STATE \textbf{Input}: A cutting-plane termination threshold $\epsilon_{cp}$.

\STATE \textbf{Initialization}: Initialize a working set $\mathcal{W} \leftarrow \emptyset$ and a random $\overline{\bfy}$;

\REPEAT

\STATE $\mathcal{W} \leftarrow \mathcal{W} \cap \overline{\bfy}$;

\STATE Obtain a solution  $\bfw$ by solving
\begin{equation}
\begin{aligned}
\underset{\bfw,\xi}{\min}
~&
\left \{
\frac{1}{2} \| \bfw \|_2^2 + C \xi \right \}\\
s.t.~&
\forall \overline{\bfy}^k \in \mathcal{W}: \\
&
\frac{1}{l} \sum_{i=1}^l \left [
\left(f^S(\bfx_i,y_i) + \bfw^\top \Psi(\bfx_i, y_i)\right) -
\left(f^S(\bfx_i,\overline{y}^k_i) + \bfw^\top \Psi(\bfx_i, \overline{y}^k_i)
\right)
\right ]
\geq
\frac{1}{l} \sum_{i=1}^l L (y_i,\overline{y}^k_i)
-\xi,
\xi \geq 0.
\end{aligned}
\end{equation}

\STATE Solve a maximization to the most violated
constraint for every $\bfx_i|_{i=1}^n$,
\begin{equation}
\begin{aligned}
\overline{y}_i = \underset{\widetilde{y}\in \mathcal{Y}}{\arg\max}
~&
\left \{
L(y_i,\widetilde{y}_i) - \left ( f^S(\bfx_i,\widetilde{y}_i) + \bfw^\top \Psi(\bfx_i, \widetilde{y}_i) \right )
 \right \}
\end{aligned}
\end{equation}

\UNTIL{$\frac{1}{l} \sum_{i=1}^l \left [
\left(f^S(\bfx_i,y_i) + \bfw^\top \Psi(\bfx_i, y_i)\right) -
\left(f^S(\bfx_i,\overline{y}^k_i) + \bfw^\top \Psi(\bfx_i, \overline{y}^k_i)
\right)
\right ]
\geq
\frac{1}{l} \sum_{i=1}^l L (y_i,\overline{y}_i)
-\xi - \epsilon_{cp}$}

\STATE \textbf{Output}: $\bfw$.

\end{algorithmic}
\end{algorithm}


\section{Conclusion}
\label{sec:conclusion}

In this paper, we extend the domain transfer learning problem to predict structured output, and propose the domain transfer structured output learning. Moreover, we also develop an algorithm to solve this problem by transferring an auxiliary structured output predictor learned from the source domain to a target domain. The transformation is implemented by adding a delta function to the auxiliary predictor so that it can be adapted to the target domain. We use the target domain samples with corresponding outputs to learn the delta function parameter via cutting-plane algorithm.

\small{

}

\end{document}